\documentclass[sigconf]{acmart}
\usepackage{enumitem}
\usepackage{balance}
\usepackage{graphicx}
\usepackage{dirtytalk}

\usepackage{fancyhdr}
\setcopyright{none}                        
\acmPrice{}                                 
\acmISBN{}                                  
\acmDOI{}                                   
\renewcommand\footnotetextcopyrightpermission[1]{}  
\setcopyright{none}                         
\begin{document}

\fancyhf{}                
\renewcommand{\headrulewidth}{0pt}  
\fancyhead[L]{}  


\title{Reward-Driven Interaction: Enhancing Proactive Dialogue Agents through User Satisfaction Prediction}

\author{Wei Shen}
\email{shenwei0917@126.com}
\authornote{These authors contributed equally to this research.}
\affiliation{%
  \institution{Baidu Inc.}
  \country{China}
}

\author{Xiaonan He}
\email{xiaonan.cs@gmail.com}
\authornotemark[1]
\authornote{Corresponding author.}
\affiliation{%
  \institution{Baidu Inc.}
  \country{China}
}

\author{Chuheng Zhang}
\email{zhangchuheng123@live.com}
\affiliation{%
  \institution{Tsinghua University}
  \country{China}
}

\author{Xuyun Zhang}
\email{xuyun.zhang@mq.edu.au}
\affiliation{
  \institution{Macquarie University}
  \city{Sydney}
  \country{Australia}
}

\author{Xiaolong Xu}
\email{xlxu@ieee.org}
\affiliation{%
  \institution{Nanjing University of Information Science and Technology}
  \country{China}
}

\author{Wanchun Dou}
\email{douwc@nju.com}
\affiliation{%
  \institution{Nanjing University}
  \country{China}
}


\begin{abstract}

    Reward-driven proactive dialogue agents require precise estimation of user satisfaction as an intrinsic reward signal to determine optimal interaction strategies. Specifically, this framework triggers clarification questions when detecting potential user dissatisfaction during interactions in the industrial dialogue system. Traditional works typically rely on training a neural network model based on weak labels which are generated by a simple model trained on user actions after current turn. However, existing methods suffer from two critical limitations in real-world scenarios: (1) Noisy Reward Supervision, dependence on weak labels derived from post-hoc user actions introduces bias, particularly failing to capture satisfaction signals in ASR-error-induced utterances; (2) Long-Tail Feedback Sparsity, the power-law distribution of user queries causes reward prediction accuracy to drop in low-frequency domains. The noise in the weak labels and a power-law distribution of user utterances results in that the model is hard to learn good representation of user utterances and sessions. To address these limitations, we propose two auxiliary tasks to improve the representation learning of user utterances and sessions that enhance user satisfaction prediction. The first one is a contrastive self-supervised learning task, which helps the model learn the representation of rare user utterances and identify ASR errors. The second one is a domain-intent classification task, which aids the model in learning the representation of user sessions from long-tailed domains and improving the model's performance on such domains. The proposed method is evaluated on DuerOS, demonstrating significant improvements in the accuracy of error recognition on rare user utterances and long-tailed domains. 
\end{abstract}

\maketitle
\section{Introduction}
Modern commercial dialogue systems like DuerOS (Baidu's voice assistant) employ modular pipelines where user satisfaction operates as the central reward signal driving proactive interactions. The primary function of a dialogue system is to understand and execute user commands through a pipeline of modules. These modules include automatic speech recognition (ASR) \cite{huang2019study, golan2018deep, bai2022joint}, rewrite query (RQ) \cite{ponnusamy2020feedback}, language understanding (LU) \cite{zhang2019joint}, information retrieval (IR) \cite{kasemsap2017mastering}, natural language generation (NLG) \cite{buck2018ask, li2017end, wen2016network}, dialogue management (DM) \cite{nikola2017neural, kuma2019practical, chen2019seman}, and text-to-speech (TTS) \cite{peng2019parallel}. Similar to other conversational systems like Alexa \cite{alok2020design} and Google Assistant \cite{tulshan2019survey}, DuerOS incorporates a proactive interaction mechanism in the DM module \cite{shen2022transformer}. The proactive interaction mechanism prompts the system to ask clarification questions when a user encounters friction while interacting with the system. Frictions can occur due to various reasons, such as ASR errors (i.e., misrecognition of user utterances), NLU errors (i.e., misunderstanding user intent), and user errors (i.e., misstating commands). Fixing these frictions with proactive interaction mechanism can help users to have a more seamless experience, and engage more with DuerOS.

Traditional approaches in conversational AI predominantly employ user satisfaction estimation models operating at either turn-level granularity or session-level aggregation\cite{bodigutla2020joint}. By using turn-level user satisfaction as the weak label and user utterances and actions prior to the current turn as features, researchers were able to train a model to predict user satisfaction with candidate responses before delivering them to the user \cite{shen2022transformer}. Based on the model's prediction, the system could determine whether or not to ask a clarification question to the user. In DuerOS,  we adopt a transformer-based model to do the user satisfaction prediction and help the system to proactively ask for clarification. Although the online experiments have demonstrated the effectiveness of the model, we try to improve the model performance and advance user experience of the system. 

Instead of changing model architecture and applying some new features to the model input, we first do some analysis of the online cases and try to find some insights to improve our model architecture. Specifically, 
we have two observations on our case analysis:

\begin{itemize}[leftmargin=*]
    \item \textbf{The model has bad performance on some rare seen user utterances.} We find that some ASR errors among online cases cannot be recognized by the model, but the user utterance of this case is implicit and the decoded text from ASR  module is rare \say{seen} in the training set. In this paper, we call these cases as rare user utterances. In addition, we find that weak label generator can generate the true label of these cases. In our view, these cases should be easily recognized by our online model due to its speciality of the decoded text. However, the model has a wrong online prediction.
    \item \textbf{The model has bad performance on some long-tailed domains.}
    We find that the model can perform well on the main domains e.g, media domain, but performs bad on some long-tailed domain e.g., Universal QA domain. Although some patterns that reveal user would not be satisfied with the candidate response can easily be found by human from training dataset, e.g, a user repeats a same utterance multiple times, the model cannot discover these patterns and recognize these user dissatisfied utterances online.   
\end{itemize}
Furthermore, we attribute these two issues to the noisy labels and a power-law distribution of the user utterances \cite{yao2021self}, which prevent the model from learning good representations of rare user utterance and user sessions in long-tailed domains. These issues may be common and non-trivial when a researcher trys to improve the model performance in proactive interaction mechanism on a industrial dialogue system.   

To address the aforementioned issues, we have implemented a multi-task learning approach to enhance the representation learning of user utterance and sessions. Specifically, we have designed two auxiliary tasks to help our model learn representations of user sessions. First, we have incorporated a contrastive learning method into our model to learn representations of rare user utterances that result from ASR errors. By using this contrastive learning method, our model can learn from the intrinsic structure of raw data and recognize rare user utterances from raw features. Second, we have included a domain-intent classification task in our model to assist in extracting some common errors in long-tailed domains. With domain-intent classification, our model can develop an improved representation of user sessions in long-tailed domains, making it easier to identify common errors in these domains. Instead of changing model architecture and add some new model inputs, this method is easily incorporated into any online model in a large dialogue system and improve the user satisfaction prediction.

To this end, this paper propose an auxiliary task based user satisfaction prediction model to learn a good representation of user sessions, especially with noisy labels and a power-law data distribution. These two auxiliary tasks can help model recognize more rare user utterances that result from ASR errors and some common errors in long-tailed domains. 

To summarize, our contribution is three-fold:
\begin{itemize}[leftmargin=*]
    \item \textbf{Revealing Common Issues in Industrial Dialogue Systems.} We do a case analysis on a large scale dialogue system and find that model trained by the real industrial datasets cannot recognize the rare user utterances that result from ASR errors and some common errors in long-tailed domains. These issues may be common when a researcher try to improve user satisfaction prediction in industrial dialogue systems.
    \item \textbf{Proposing two auxiliary tasks to improve user satisfaction prediction}
    To solve the above mentioned issues, we propose two auxiliary tasks to enhance the model representation learning. Specifically, we incorporate a contrastive learning task into our base model that enhance the representation learning of the user utterances, especially on those rare seen user utterances. Furthermore we incorporate a domain-intent classification task into our base model that enhance the representation learning of user sessions especially on long-tailed domains and help the model discover more common errors in some long-tailed domains. 
    \item \textbf{Offline Experiments and Online Experiment On a Industrial Dialogue System.}
    The offline experiments on our industrial dataset demonstrate the effectiveness of our method. Furthermore, we conduct an online evaluation by deploying the model to DuerOS, which achieve a significant improvement on a session-level user satisfaction evaluation. With further analysis of the online experiment, our model truly improves the model performance on the accuracy of error recognition on rare user utterances and long-tailed domains.
\end{itemize}

\section{Related Work}
Our work touches on three strands of research: proactive interaction mechanism in spoken dialogue system, contrastive self supervised learning method and multi-task learning method

\subsection{Proactive interaction mechanism in spoken dialogue system}
Proactive interaction mechanism is a well-researched topic in the field of spoken dialogue systems, and it involves two key challenge tasks: when to ask for clarification in each turn, and how to generate clarification queries \cite{majumder2021ask, severcan2020clarification, zamani2020mimics}. In our paper, we focus specifically on when to ask for clarification in each turn. Previous research by Alok et al. \cite{alok2020design} proposed a hypothesis rejection module that uses a deep model with the user utterance and a series of handcrafted features from the NLU module to predict whether the system should reject the NLU result or provide a response to the user in Alexa. If the NLU result is rejected, Alexa may ask for clarification or provide no response to the user. However, previous models have only focused on the information available in the current turn, ignoring the rich information that may be across the user historical sessions, which could reveal the user's preferences and potential intentions, such as whether the user prefers a short video or not. 

In addition, Shen et al. propose a transformer-based model \cite{shen2022transformer} to do user satisfaction prediction in DuerOS. Specifically, they first generate a large number of weak labels according to the user’s interactions with the system in the current turn and the next turn. Based on these weak labels, they propose a transformer-based model to extract information from both the structured and text data and grasp the temporal dependency between the current turn and the previous turns for user satisfaction prediction. Depending on the predicted satisfaction level, the DM module decides whether to provide the candidate response or ask the user for clarification. However, when the model is deployed to DuerOS and we do a further analysis of the user online sessions, the transformer-based model has a bad performance on the recognition of some rare user utterances that result from ASR errors and some common errors occurring on long-tailed domains. As a result, we propose two auxiliary tasks to improve user dissatisfaction prediction.    

\subsection{Contrastive self supervised learning method}
The concept of Contrastive Self-Supervised Learning (SSL) has garnered significant attention from various research communities such as computer vision (CV), natural language processing (NLP), recommendation systems and reinforcement learning \cite{guo2022contrastive, tao2022self, yang2022reading, cai2022imitation, cai2022td3}. The primary objective of contrastive SSL is to enhance the discrimination ability of negative data while maximizing mutual information among positive data transformations. For instance, in the context of recommendation systems, researchers have proposed a two-tower Deep Neural Network (DNN) based contrastive SSL model to improve collaborative filtering by utilizing item attributes. 

In the NLP area, SimCSE (Similarity-based Contrastive Self-Supervised Learning for Natural Language Processing) is a state-of-the-art approach for learning general-purpose sentence representations through self-supervised learning \cite{gao2021simcse}. It maximizes the similarity between augmented versions of the same sentence while minimizing the similarity between augmented versions of different sentences. The main idea behind SimCSE is to train a neural network to encode the semantic content of sentences into low-dimensional vector representations that can be used for various downstream NLP tasks such as text classification, information retrieval, and question answering. In our paper, we adopt SimCSE method to help model learn the representation of user utterances. It serves as a auxiliary task and can assist model in recognizing rare user utterances that may result from ASR model.

\subsection{Multi-task learning method}
Multi-task learning (MTL) is a machine learning technique that enables a model to learn to perform multiple related tasks simultaneously \cite{kollias2023abaw, kurin2022defense, castellucci2019multi, zhao2019recommending, shen2020auxiliary,li2021fair,liu2025adaptivestep,cai2023reprem}. In traditional machine learning, each task is trained separately, which can be inefficient, especially when there is a common underlying structure shared by multiple tasks. With multi-task learning, a single model is trained to perform multiple tasks at the same time, sharing the common knowledge and structure among them. This can lead to better generalization and improved performance on all tasks.

Multi-task learning is often used in natural language processing (NLP), computer vision, and speech recognition tasks, where multiple related tasks can benefit from shared representations, such as shared word embeddings in NLP or shared visual features in computer vision. In our paper, we incorporate contrastive self supervised learning task and domain-intent classification task into our model, which enhance the learning of the representation of user utterances and sessions. We consider that some patterns in the long-tailed user utterances and user sessions can be recognized under the multi-task training and generalized well in the user satisfaction prediction. Though the model pretrained by these two auxiliary tasks can get similar or better performance on user satisfaction prediction, the process of pretraining model is time-consuming and therefore not suitable for an industrial dialogue system. 

\section{System Overview}
In this section, we will briefly introduce the proactive interaction mechanism, the architecture of the base model in the mechanism and how the model serves for the system.
\subsection{The proactive interaction mechanism}
The proactive interaction mechanism is to prompts the system to ask for clarification when a user encounters friction when interacting with a industrial dialogue system. With this mechanism, a industrial dialogue system always follows these steps to generate a response based on a user's utterance (voice command):
\begin{itemize}[leftmargin=*]
    \item The ASR module converts the utterance into a textual query and the RQ module may rewrite this textual query based on a user historical sessions.
    \item The LU module identifies the domain and intent of the query, and fills in any associated slots in the domain-specific semantic template.
    \item The IR and NLG modules work together to generate a candidate response, which consists of a suitable result item and a voice response, based on the domain-intent and slots provided. The voice response is usually a piece of audio from the TTS module. For simplicity, we use voice response to refer the input text of the TTS module.
    \item The proactive interaction mechanism in DM module uses the candidate response to predict user satisfaction. Depending on the predicted satisfaction level, the DM module decides whether to provide the candidate response or ask the user for clarification.
\end{itemize}

\begin{figure*}[!t]
 	\centering\includegraphics[width=7.0in]{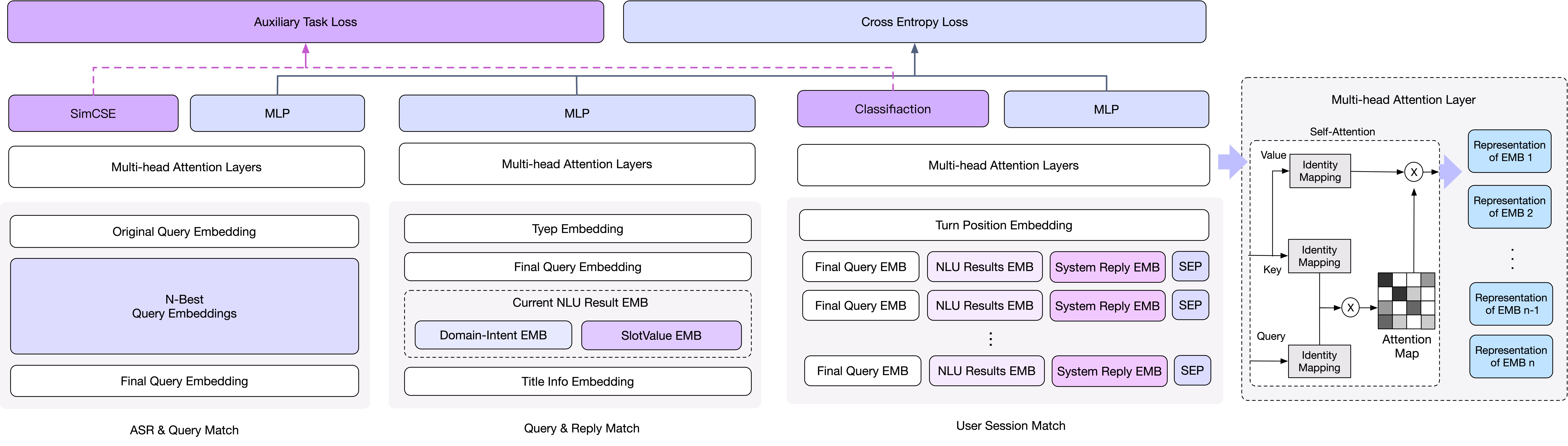}
 	\caption{An overview of our model. The model contains three parts: 1) Left part of the model, the ASR and query match module, is to model the match between user's current voice and ASR decoded texts that contains the original query, n-best queries from the ASR module, and the final query from the rewrite module. 2) Middle part of the model, the query and reply match module, is to model the match between user utterances and system results that contains the final query, NLU results, and title info from the IR module.
    3) Right part of the model, the user session match module, is to model the match among user recent turns, which contains multi-turn user utterances and system information (system response, the intervals between two turns, and the system NLU results).
 	}\label{fig:overview}
 \end{figure*}
 
\subsection{The base model in the proactive interaction mechanism}
In the proactive interaction mechanism, we train a transformer-based model to make user satisfaction prediction. 
In this model, we have three parts, the ASR and query match module, the query and reply match module, and the user session match module.

In the ASR and query match module, we use original query $q_o$, n-best queries from ASR module $q_n$ that are top-k texts decoded from a user utterance by ASR module, and final query $q_f$ from the RQ module as the inputs, where each word in a query is encoded to a embedding (we denote them as $e_o$, $e_n$ and $e_f$) by the embedding layer and therefore each text is encoded as sentence embedding $s_e$. 
\begin{equation*}
    e_o, e_n, e_f = EMB(q_o, q_n, q_f)
\end{equation*}
\begin{equation*}
    s_e = Concate(e_o, e_n, e_f)
\end{equation*}

Then we adopt multiple multi-head attention layers to do the interactions between these multiple input texts. At last, we adopt reduce-max operation to extract the \emph{error} ($T_q$) from these vectors.
\begin{equation*}
    o_e = MultiHeadAtt(s_e, N)
\end{equation*}
\begin{equation*}
    T_q = ReduceMax(o_e)
\end{equation*}
With the help of multi-head attention layers, we hope the interactions between multiple queries can assist model in recognizing the mismatch between the queries and discovering some ASR errors. For example, if the decoded queries are not similar, there may exist errors in the ASR module.

In the query and reply match module, we use the final query $q_f$ generated by the RQ module, the domain-intent and slot information $h$ extracted by the NLU module, and the textual description of the item retrieved by the IR module. The query and textual description $r$ are encoded into sentence vectors $e_r$ using the embedding module. The NLU results are also mapped into IDs and then encoded into vectors $e_h$. 
\begin{equation*}
    e_q, e_h, e_r = EMB(q_f, h, r)
\end{equation*}
\begin{equation*}
    e_p = Concate(e_q, e_h, e_t)
\end{equation*}
The model concatenates all these vectors into a single vector $r_e$ and sequentially applies $N$ multi-head attention layers and reduce max operator, which output the \emph{error} vector $T_r$. 
\begin{equation*}
    o_p = MultiHeadAtt(e_p, N)
\end{equation*}
\begin{equation*}
    T_r = ReduceMax(o_p)
\end{equation*}
This design allows the model to capture the interactions between the final query, NLU results, and textual descriptions and can identify the mismatch between the user utterance and system response. For instance, if the system decodes the user utterance into give me a music \emph{show up} but the IR module retrieves a short video \emph{show up}, the model can easily recognize the mismatch between \emph{music} in the user utterance and \emph{short video} in the IR result.

In the user session match module, we use the user sessions $\hat{S}$ as the input where the $i$-th turn contains final query $q_f^i$, NLU results (domain, intent and slots) $n^i$ and time interval between the current utterance and the former utterance $t^i$. Similarly, each query is encoded to a sentence embedding $e_q^i$. each domain-intent and slot are mapped into a unique NLU embedding $e_n^i$. Besides, the time interval are discretized and indexed into an id, then embedded into a embedding $e_t^i$. At last, the model concatenate these vectors and a special embedding (SEP or $e_s^i$) as one vector $s_i$, which represents a user turn. 
\begin{equation*}
    e_q^i, e_n^i, e_t^i = EMB(q_f^i, n^i, t^i)
\end{equation*}
\begin{equation*}
    e_s^i = Concate(e_q^i, e_n^i, e_t^i)
\end{equation*}
With multiple user turn representations, we adopt the $N$ multi-head attention layers to do the interactions among user turns in a user session. We hope these interactions may contain the error patterns among the user turns. For example, if a user inputs the same utterances multiple times in a short time, the user may be not satisfied with the system results. At last, we adopt reduce-max operation to recognize the \emph{error} $T_s$ in the user sessions.
\begin{equation*}
    E_s = Concat(e_s^0, e_s^1, ..., e_s^L)
\end{equation*}
\begin{equation*}
    o_s = MultiHeadAtt(E_{s}, N)
\end{equation*}
\begin{equation*}
    T_s = ReduceMax(o_s)
\end{equation*}

In the final layer, we concatenate three vectors ($T_q, T_r, T_s$) into a vector and then use a fully-connected layer to encode the vector into a single value. By the sigmoid function, the model outputs the probability of user satisfaction. In addition, we use the cross entropy ($CE$) between the prediction ($p$) and the labels ($l$) as the loss function.
\begin{equation*}
    \mathscr{L}_{main} = CE(p, l)
\end{equation*}

\begin{figure}[!t]
 	\centering\includegraphics[width=3.0in]{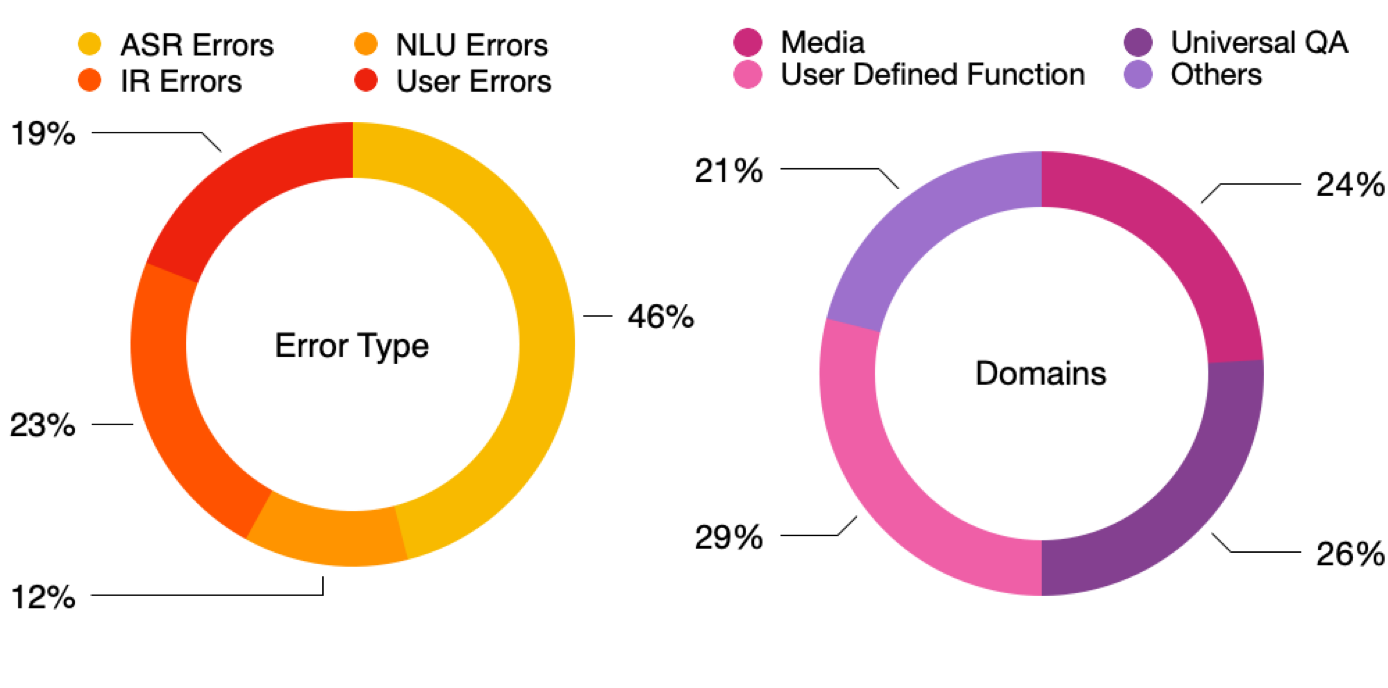}
 	\caption{The analysis of the online error cases. Among the unrecognized 230 user dissatisfied cases, the ASR errors are the main error type and occupy 48\% of total 220 cases. Furthermore, we find that some long-tailed domains e.g, Universal QA, occupy a similar account of undiscovered errors with those on the main domain, i.e., media domain}
    \label{fig:case-analysis}
 \end{figure}
 \subsection{How to deploy model to the DM module and serve for the system ?}

 When we deploy the model into the DM module, the model may increase the system latency and bring the user unsatisfactory. To reduce the system latency, we split the main model into three sub-modules. This architecture is easily deployed into our system. We can implement the model prediction in the different stages: we first do the model prediction of the ASR and query match module after the ASR module, then do the model prediction of the user session match module after the LU module, and do the model prediction of the query reply match module after the IR module. At last, we combine these sub model prediction into the final prediction in the DM module. As a result, the first two predictions of the sub modules can be implemented parallel with other system modules. The model only adds the time consumption of the query reply match module into the system latency. In addition, we hope three sub-module can recognize different type of erros.
 
At last, we will set a threshold that help model recall the user dissatisfied utterance. if the model prediction is smaller than the threshold, the system will actively propose a clarification query to the user. The threshold is selected by the offline experiments.  

\section{Auxiliary tasks training process}
In this section, we will show the analysis of the online performance of transformer-based model and propose what motivates us to design these two auxiliary tasks to improve user satisfaction prediction. Then we specify the two auxiliary tasks, contrastive self-supervised learning task and domain-intent classification task. At last, we briefly show how to incorporate these two tasks into our model.

\subsection{The analysis of the transformer-based model prediction}
When we deployed the base model to our system for one month. We extract $1000$ user sessions uniformly in this month and analyze the model prediction of our model. In these cases, there are about 670 cases that users are satisfied with the system results. In the other 330 cases, the base model can recognize about 100 case of them that users are not satisfied with the system result. As showed in Figure \ref{fig:case-analysis}, we do the further analysis on the left 230 cases and find that the model cannot discover a large amount of ASR errors and errors in the long-tailed domains. 

In these left errors, there are a large number of errors that results from ASR module or user errors that user utterances do not give a explicit intent. Furthermore, we find that these user utterances always consists of some rare words and have not explicit meanings. Additionally, we analyze the error distributions among different domains. In our system, the media domain is the main domain which user utterances are mainly belong to. However, there are only 24 percent left errors in this domain. It indicates that the online model performs well on the main domain. However, there are about 76 percent left errors on universal QA, user defined function (UDF) and other domains. It indicates that our model performs badly on these long-tailed domains

With the case analysis of the online model, we find that the model does not perform well in recognizing some ASR errors and errors on long-tailed domains.  

\subsection{Auxiliary tasks}
Based on the above analysis, we design two auxiliary tasks to assist the model in improving user satisfaction prediction. 

\begin{figure*}[!t]
 	\centering\includegraphics[width=6.5in]{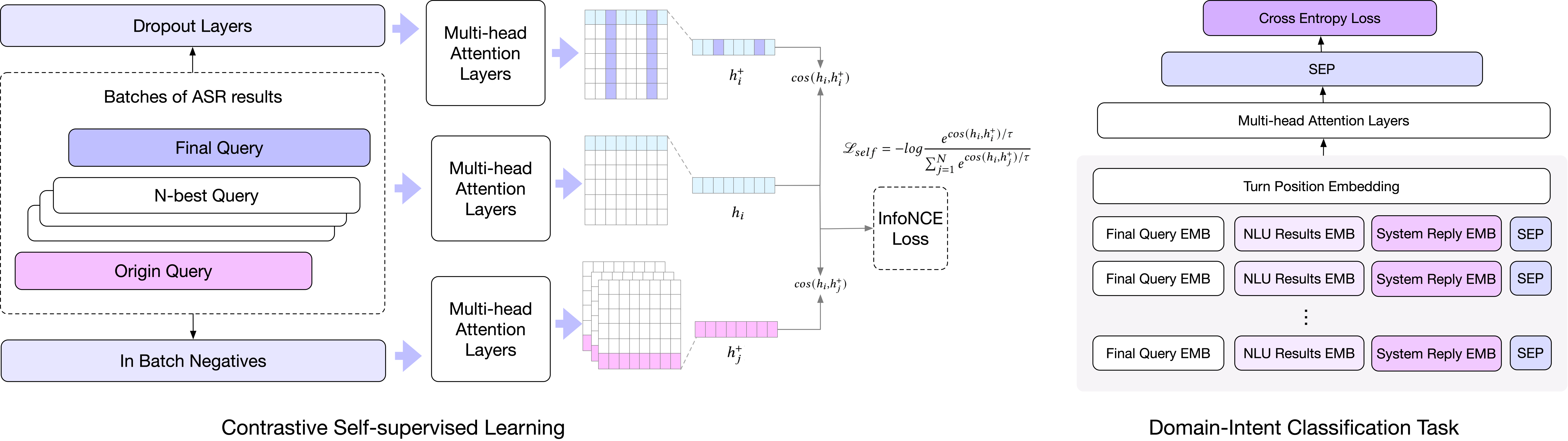}
 	\caption{Two auxiliary tasks to enhance the representation learning of the user utterances and user sessions. The left part is a contrastive self-supervised learning task: we pass the same user queries to the embedding layer twice and apply the standard dropout operator to the duplicated vectors of queries before multiple multi-head attention layers. Then we can obtain two different embeddings as positive pairs, take other queries in the same mini-batch as negatives and the auxiliary task is to predict the positive one among negatives. The right part is the domain-intent classification task: we first extract the representation of the user session, pass the vector of the symbol \emph{SEP} in the current turn to a fully connected layer, and adopt cross-entropy loss to train the classification task.
 	}\label{fig:ssl}
 \end{figure*}
 
\subsubsection{Contrastive self-supervised learning task}
The model always performs badly on a large amount of ASR errors and implicit user utterances. Furthermore, we find that these user utterances are not easy to be merged into a cluster in the training process. It may result from insufficient user unsatisfied samples. Accordingly, we adopt a contrastive self-supervised learning (SSL) method to track the sample sparsity problem by learning better latent relationships of user utterances and sessions. Specifically, we adopt the SimCSE method, a state of art NLP self-supervised learning method to help train user utterance embedding.

As shown in the left part of Figure. \ref{fig:ssl}, we apply the contrastive SSL method for training the original query embedding, n-best queries embeddings, and final query embedding. Without any data augmentation method, we pass the same user queries to the embedding layer twice and apply the standard dropout operator to the duplicated vectors of queries before multiple multi-head attention layers. Then we can obtain two different embeddings as positive pairs, and take other queries in the same mini-batch as negatives. Given the vectors of user queries, we adopt multiple multi-head attention layers to generate the query embeddings. In addition, we share the model parameters to model each type queries (i.e., original query, n-best queries, and final query) to improve the model generalization. At last, we take a cross-entropy objectives with in-batch negatives: let $h_i$ and $h_i^+$ denote the presentations of $q_i$ and $q_i^+$, the training objective for $(h_i, h_i^+)$ with mini-batch of $N$ pairs is:
\begin{equation*}
    \mathscr{L}_{self} = -log\frac{e^{cos(h_i, h_i^+) / \tau}}{\sum_{j=1}^{N}e^{cos(h_i, h_j^+)/ \tau}}
\end{equation*}

\subsection{Domain-intent classification task}
The user utterances are often highly skewed in a power-law distribution \cite{yao2021self}. Accordingly, this will always leave the training data for long-tail domains very sparse and result in bad model performance on the user utterances in long-tailed domains.

To enhance the model performance on the long-tailed domain, we adopt a domain-intent classification task to cluster the representations of user sessions. As shown in the right part of Figure .\ref{fig:ssl}, we first extract the representation of the user session, pass the vector of the symbol \emph{SEP} in the current turn to a fully connected layer and adopt cross-entropy loss to train the classification task. Specifically, we extract the representation vector $s$ from the user session sub-module. After the fully connected layer and sigmoid function, the vector $s$ is transformed as a prediction $p$. Then we adopt the binary cross entropy loss function to get the final loss on $p$ and labeled domain-intent $d$:
\begin{equation*}
    \mathscr{L}_{cl} = CE(p, d)
\end{equation*}

\subsection{Incorporate the auxiliary tasks into the model} 
There are two main types of methods to use the self-supervised learning method to improve model representation learning. One is to pre-train the model with the self-supervised learning task and finetune the model with the user satisfaction prediction task based on the pre-train model. The other is to use the self-supervised learning task as an auxiliary task to enhance the representation learning during the training user satisfaction prediction task. However, the first method consumes too much computing resources and is time-consuming, which is not suitable for daily training. Accordingly, we adopt the second method to incorporate two auxiliary tasks into the base model. Specifically, we use a weighted sum of the user satisfaction prediction task loss, contrastive self-supervised learning task loss, and domain-intent classification task loss as the final loss:
\begin{equation*}
    \mathscr{L} = \mathcal{L}_{main} + w_1\mathcal{L}_{self} + w_2\mathcal{L}_{cl}
\end{equation*}

\section{Experiment}
In this section, we conduct experiments on one real dataset and evaluate our model online on DuerOS. We aim to answer the following questions:

\textbf{Q1:} How does the online model perform compared with other baseline models?

\textbf{Q2:} How does the auxiliary task-based model perform compared with the online model?

\textbf{Q3:} How do two auxiliary tasks influence the performance of our model respectively?

\textbf{Q4:} How does our auxiliary task-based model perform when we deploy it on DuerOS?
\subsection{Offline Experiments}
\subsubsection{Dataset} We conduct experiments on an industrial dataset collected from DuerOS. The training and validation data of the dataset are extracted from a month. The training data contains 32M samples. The validation data contains 250K samples. The training data and the validation data are all labeled by the weak label generator. The testing data contains 1k samples and are labeled by expert annotators, which can be regarded as the ground truth.

\subsubsection{Metrics and hyperparameters} In our experiment, we use the area under the curve (AUC) and the conditional label accuracy (CLA) as the evaluation metrics. AUC is the area under the ROC curve, and CLA is the maximum recall when precision is larger than a specified value (which is set to 85\% in this paper). In our experiments, we binary our prediction according to a threshold and then calculate the classification precision given the true labels, where we tune the threshold by grid searching to maximize the recall rate of dissatisfied user sessions. 

 We implement our model based on Tensorflow and use the Adam optimizer. We apply a grid search for hyperparameters of the based model: the number of layers in two transformer blocks in each sub-module is searched in $\{1, 2, ..., 8 \}$ and the embedding size of each vector in the embedding layer is chosen from $\{90, 120, ..., 1800\}$. Besides, the number of turns in the user session sub-module is chosen from $\{1, 2, ..., 10\}$. (cf. Figure \ref{fig:overview}). At last, we apply a grid search for hyperparameters of auxiliary task based model: the weights of the loss of two auxiliary tasks are choosen from $\{10^{-1}, 10^{-2}, 10^{-3}\}$. Then the weight of the contrastive self supervised task is $10^{-2}$ and that of the domain-intent classification is $10^{-1}$. The selected hyperparameters of the model are listed in the Tabel \ref{table:Hyp}.

 \begin{table}[!h]
    \caption{Hyperparameters in the model}
	\centering
	\begin{tabular}{lr}
       	\hline
	    Hyperparameters & Value\\
		\hline
 LR & $1.2\times 10^{-3}$ \\
 The number of turns & $5$ \\
 Embedding size & 320\\
 Threshold & 0.78\\
 \#transformer layers in first sub-module & $4$ \\
 \#transformer layers in second sub-module & $4$ \\
 \#transformer layers in third sub-module & $4$ \\
 The weight of the CSSL task & $10^{-2}$ \\
 The weight of the domain-intent classification task & $10^{-1}$ \\
		\hline
	\end{tabular}
	\label{table:Hyp}
\end{table}

\begin{table*}[t]
	\centering
	\caption{The performance comparison between the baselines and our model on testing sets.}
	\begin{tabular}{l|ccccc}
		\hline
		Model & Dataset (AUC / CLA) & Media (AUC / CLA) & Univesal QA (AUC / CLA) & UDF (AUC / CLA) & Others (AUC/CLA)\\
		\hline
		LSTM & 0.765 / 0.060 & 0.772 / 0.070& 0.640 / 0.030& 0.720 / 0.054& 0.710 / 0.052\\ 
		CNN &  0.734 / 0.051 & 0.761 / 0.062& 0.600 / 0.020& 0.693 / 0.040& 0.682 / 0.038\\
		TDU & 0.771 / 0.071 & 0.780 / 0.078& 0.645 / 0.038& 0.722 / 0.051& 0.681 / 0.037\\
        TBM & 0.782 / 0.082 & 0.791 / 0.085& 0.652 / 0.040& 0.725 / 0.055& 0.680 / 0.033\\
        TBM-2 & \textbf{0.792} / 0.090 & \textbf{0.795} / 0.093& \textbf{0.670} / 0.045& 0.722 / 0.052 & \textbf{0.685} / 0.040\\
        ABM-S  & 0.789 / 0.091 & \ 0.793 / 0.097 & 0.672 / 0.043 &  0.725 / 0.050 & 0.682 / 0.042\\
        ABM-C  & 0.788 / 0.089 & \ 0.790 / 0.090 & 0.665 / 0.055 &  0.723 / 0.058 & 0.684 / 0.050\\
		ABM (ours)  & 0.790 / \textbf{0.095} & \ 0.793 / \textbf{0.098} & 0.667 / \textbf{0.058}& \textbf{0.735} / \textbf{0.055} & 0.675 / \textbf{0.058}\\
		\hline
	\end{tabular}
	\label{table:performance_compare}
\end{table*} 

 \subsubsection{Baselines} We compare our auxiliary task-based model (ATM) with the following baselines:
 \begin{itemize}[leftmargin=*]
    \item \textbf{TDU.} TDU \cite{zhang2019joint} is a BiLSTM based deep neural net model in Alexa that predicts both turn-level and dialogue-level user satisfaction.
    \item \textbf{TBM}
     TBM is a former transformer-based deep neural net model which has been deployed to the DuerOS.
    \item \textbf{TBM-2}
     TBM-2 is the new transformer-based deep neural net model which consists of three sub-modules to model the ASR and query match, user session match and query reply match respectively.
     \item \textbf{CNN and LSTM} We also implement several popular architectures: CNN (following Text-CNN in Kim \citet{kim2014convolutional}) and LSTM \cite{hochreiter1997long}. We train the models with these architectures using the same training sets.
\end{itemize}

\subsection{Results} In the subsequent paragraphs, we first present the performance comparison between the baselines and the auxiliary task-based model on the datasets, and an ablation study of our model. At last, we give a further analysis of how two auxiliary tasks improve the user satisfaction prediction. 

\textbf{Performance comparison (Q1 \& Q2).} We show the experimental results on the  offline dataset in Table \ref{table:performance_compare}. We have the following observations on the experimental results:
\begin{itemize}[leftmargin=*]
    \item Our auxiliary task-based model (ABM) outperforms all the other baselines on the metric CLA. In particular, compared with TBM-2, the performance of our model has improved  5.5\% in CLA on the offline dataset. Specifically, it gains 5.3\% CLA improvement in the media domain, 28.8\% CLA improvement in the universal QA domain, 5.7\% CLA improvement in the UDF domain, and 45 \% CLA improvement in other domains. Our ABM model gains limited CLA improvement on the media domain which is the main domain in the system, but gains large CLA improvement on the universal QA and some long-tailed domains. It indicates that two auxiliary tasks help the model learn better user session representations on these domains.      
    \item The TBM-2 and ABM outperforms all the other baselines on the all domains in the experiments. It indicates that our three sub-modules design improve the model user satisfaction prediction. Though this design result in that a series of ASR results (original query and n-best queries), user sessions and system results have no interactions, the model still compete the the strong baselines (TDU and TBM) on metric AUC and CLA. It indicates that these interactions may be unnecessary in the former models. 
    \item Though the ABM model does not outperform all the other baselines on metric AUC, it gains much improvement on the CLA metric. The metric CLA is the crucial metric that instructs us on how to select a threshold online. We use this threshold to help the model select the user's potentially dissatisfied candidates and then actively ask for clarification. However, according to our offline experiments, the metric AUC is not very consistent with the metric CLA when the model can gain a high AUC on the offline dataset. In further offline experiments, we will focus on improving the model performance on metric CLA on the test dataset.
\end{itemize}

\textbf{Ablation Study (Q3).} Here, we study how two auxiliary tasks affect the model performance. Specifically, we respectively incorporate a contrastive self-supervised learning task into the model (we denote it as ABM-S) and a domain-intent classification task into the model (we denote it as ABM-C).  With compared to the TBM-2 which is the model trained without auxiliary tasks and ABM which is the model trained with two auxiliary tasks, we have the following observations:
\begin{itemize}[leftmargin=*]
    \item Compared with TBM, ABM-S has achieved the better performance on media domain and a similar performance on the long-tailed domains. It may result from the contrastive self supervised learning help model learn better user utterance and session representation.
    \item Compared with TBM, ABM-C has achieved the better performance on the long tailed domains (universal QA domain, user defined function domain and some other domains), but perform worse than the base model on the media domain. It indicates that the domain-intent classification task help model learn the better representation of the user sessions on some long-tailed domains.
    \item With two auxiliary tasks, the ABM model achieve the better performance on all domains. It indicates that these two auxiliary tasks do not conflict when training together with the main task. 
\end{itemize}

\begin{figure*}[!t]
 	\centering\includegraphics[width=5.5in]{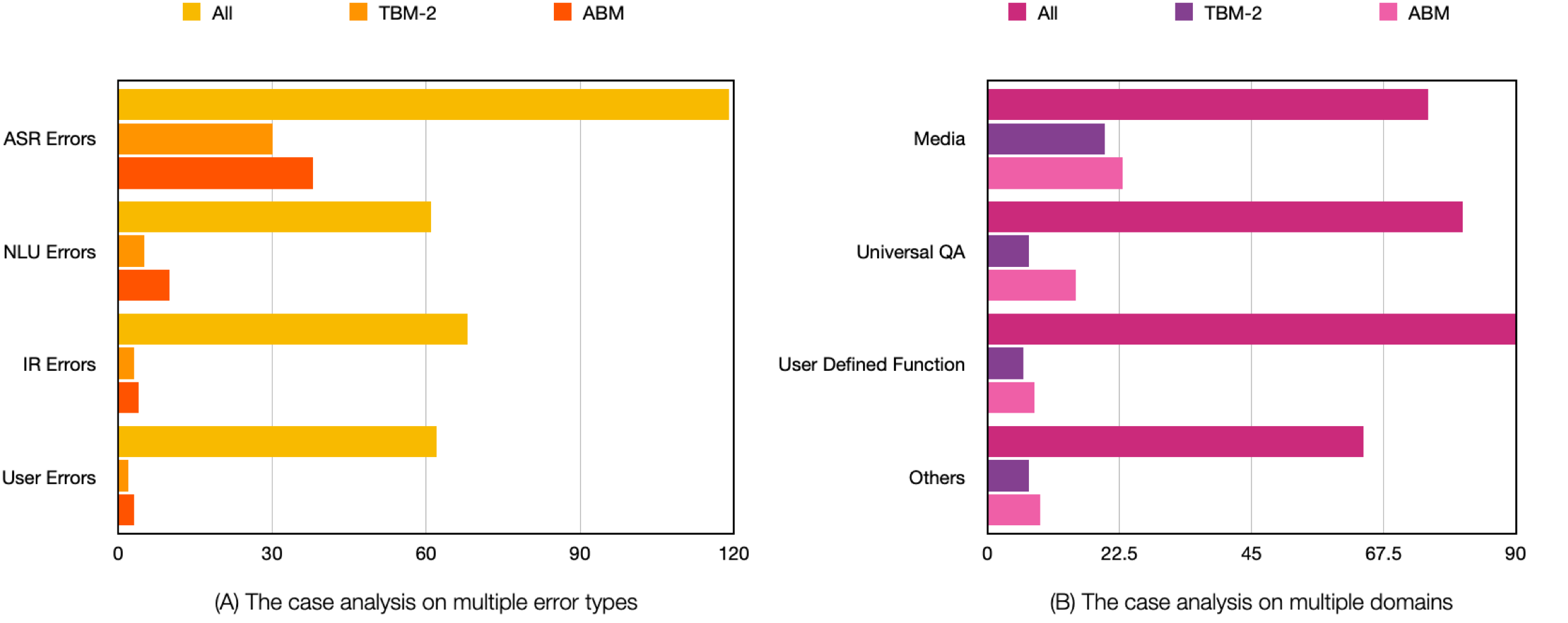}
 	\caption{The online analysis of the performance of TBM-2 and ABM. We further compare the performance of these two methods on different domains and different error types. Specifically, we filter the user dissatisfaction sessions from 1000 user sessions and divide them by domains and error types. Among 1000 online cases, there are 310 user dissatisfaction user sessions out of 1000 user sessions. The ABM and TBM-2 have a similar accuracy of error recognition (86.5\% and 85\%). However, the ABM method recalls 38 ASR errors from a total of 119 ASR errors and 10 NLU errors from a total of 61 NLU errors, which is better than TBM-2 model that recall 30 ASR errors and 5 NLU errors. In addition, the ABM method recalls 15 user-dissatisfied sessions on the universal QA domain, which is significantly better than the TBM-2 model that recall 7 sessions.}
    \label{fig:exp-analysis}
 \end{figure*}
\section{Online Experiment}
With the encouraging results on the offline dataset, we perform online experiments on DuerOS. We deploy this model to the DM module of DuerOS to control whether to ask a clarification question or directly give the candidate response to the user. Before presenting the results, we first introduce the experimental setup for the online experiments.

\textbf{Experiment setup and evaluation metrics.} We use A/B testing, which is originated from Deininger and widely used to perform controlled experiments with two or more variants in real-world systems. In our experiments, we partition the users into 50/50 groups for two variants: In the first group, we adopt the TBM-2 as the user satisfaction predictor. In the second group, we use ABM as the user satisfaction predictor. In addition, we train each model with the training samples in the last week and update them every day. All experiments lasted for two weeks to avoid daily fluctuations.

We measure the online performance for each model using contextual user satisfaction (CUS) where the label comes from expert annotators \cite{shen2022transformer}. To estimate CUS, we first sample the same number of user queries (1000 queries) from the queries in each group. Then, we extract the context for each query (e.g., 10 turns before the query, 10 turns after the query, and the candidate response), manually annotate whether it is a satisfied query for each sample, and calculate the CUS for the target turn.

\textbf{Results.} The average CUS ratings across samples for the two groups (TBM-2 and \textbf{ABM}) are 0.684 and \textbf{0.701} respectively. We see that the auxiliary task-based model achieved the best performance gain in terms of CUS, which is consistent with previous offline experiments. Given these promising results, our model has been successfully deployed to the DM module of DuesOs, serving hundreds of millions of user utterances daily.

\textbf{Further Analysis.} We further compare the performance of these two methods on different domains and different error types. Specifically, we filter the user dissatisfaction sessions from 1000 user sessions and divide them by domains and error types. As shown in Figure \ref{fig:exp-analysis}, there are 310 user dissatisfaction user sessions out of 1000 user sessions. Although ABM and TBM-2 have a similar accuracy of error recognition (86.5\% v.s. 85\%) on online cases, the ABM method recalls 38 ASR errors from a total of 119 ASR errors and 10 NLU errors from a total of 61 NLU errors, which is significantly better than the TBM-2 method. In addition, the ABM method recalls 15 user-dissatisfied sessions on the universal QA domain, which is significantly better than the TBM-2 method. With this online analysis, we find that our method truly improves the model performance on ASR errors and the long-tailed domains and help user have a more seamless and engaged experience.

\section{Conclusions}

Our study exposes a critical challenge in reward-driven proactive interaction: industrial dialogue systems struggle to accurately model user satisfaction when processing rare ASR-corrupted utterances or long-tail domain patterns. This issue stems from noisy observed states and imbalanced data distributions, which ultimately degrade the agent's ability to optimize proactive clarification strategies. We propose an auxiliary task based model to improve user satisfaction prediction in a industrial dialogue system, DuerOS. Specifically, two auxiliary tasks are designed to enhance the model's performance: The first task is a contrastive self-supervised learning method that helps the model learn the representations of rare user utterances, enabling it to recognize more ASR errors from such utterances. The second task is a domain-intent classification task that helps the model learn the representation of user sessions in long-tailed domains. By improving the representations, the model can identify common errors from long-tailed domains more accurately. We conduct experiments on a large industrial dataset and deploy the model to a industrial spoken dialogue system, evaluating the proactive interaction mechanism using a new metric called contextual user satisfaction. The results demonstrate that the proposed model recalls more errors from ASR module and long-tailed domains online and truly improve the session-level user satisfaction on DuerOS.

\bibliographystyle{ACM-Reference-Format}
\balance
\bibliography{sample-base}

\end{document}